\documentclass{article}

\usepackage[preprint]{neurips_2025}
\usepackage{enumitem}
\usepackage{amsmath}
\usepackage{amssymb}
\usepackage{graphicx}
\usepackage{adjustbox}
\usepackage{booktabs}   % 增强表格样式
\usepackage{multirow}   % 多行/列合并
\usepackage{makecell, array}
\usepackage[detect-weight]{siunitx}    % 数字对齐（可选）
\usepackage{url} % 支持URL格式
\usepackage[utf8]{inputenc} % allow utf-8 input
\usepackage[T1]{fontenc}    % use 8-bit T1 fonts
\usepackage{hyperref}       % hyperlinks
\usepackage{url}            % simple URL typesetting
\usepackage{booktabs}       % professional-quality tables
\usepackage{amsfonts}       % blackboard math symbols
\usepackage{nicefrac}       % compact symbols for 1/2, etc.
\usepackage{microtype}      % microtypography
\usepackage{xcolor}         % colors

\title{FreshRetailNet-50K: A Stockout-Annotated Censored Demand Dataset for Latent Demand Recovery and Forecasting in Fresh Retail}

\author{%
  Yangyang Wang\\
  Dingdong Ltd.\\
  \texttt{wangyangyang28@100.me} \\
  \And
  Jiawei Gu \\
  Dingdong Ltd.\\
  \texttt{gujiawei07@100.me} \\
  \And
  Li Long \\
  Dingdong Ltd.\\
  \texttt{longli01@100.me} \\
  \AND
  Xin Li \\
  Dingdong Ltd.\\
  \texttt{li.xin.cv@gmail.com} \\
  \And
  Li Shen \\
  Sun Yat-Sen University \\
  \texttt{mathshenli@gmail.com} \\
  \And
  Zhouyu Fu \\
  Unaffiliation \\
  \texttt{fu.zhouyu@gmail.com} \\
  \And
  Xiangjun Zhou \\
  Dingdong Ltd.\\
  \texttt{zhouxiangjun@100.me} \\
  \And
  Xu Jiang \\
  Dingdong Ltd.\\
  \texttt{jiangxu01@100.me} \\
}

\begin{document}

\maketitle

\begin{abstract}

 Accurate demand estimation is critical for the retail business in guiding the inventory and pricing policies of perishable products. However, it faces fundamental challenges from censored sales data during stockouts, where unobserved demand creates systemic policy biases. 
Existing datasets lack the temporal resolution and annotations needed to address this censoring effect. 
To fill this gap, we present FreshRetailNet-50K, the first large-scale benchmark for censored demand estimation. It comprises 50,000 store-product time series of detailed hourly sales data from 898 stores in 18 major cities, encompassing 863 perishable SKUs meticulously annotated for stockout events. The hourly stock status records unique to this dataset, combined with rich contextual covariates, including promotional discounts, precipitation, and temporal features, enable innovative research beyond existing solutions.  
We demonstrate one such use case of two-stage demand modeling: first, we reconstruct the latent demand during stockouts using precise hourly annotations. We then leverage the recovered demand to train robust demand forecasting models in the second stage. Experimental results show that this approach achieves a 2.73\% improvement in prediction accuracy while reducing the systematic demand underestimation from 7.37\% to near-zero bias. With unprecedented temporal granularity and comprehensive real-world information, FreshRetailNet-50K opens new research directions in demand imputation, perishable inventory optimization, and causal retail analytics. The unique annotation quality and scale of the dataset address long-standing limitations in retail AI, providing immediate solutions and a platform for future methodological innovation. The data (\href{https://huggingface.co/datasets/Dingdong-Inc/FreshRetailNet-50K}{https://huggingface.co/datasets/Dingdong-Inc/FreshRetailNet-50K}) and code (\href{https://github.com/Dingdong-Inc/frn-50k-baseline}{https://github.com/Dingdong-Inc/frn-50k-baseline}) are openly released.

\end{abstract}

\section{Introduction}
\begin{figure}[t]
    \centering
        \centering
        \includegraphics[width=0.85\textwidth, height=5.3cm]{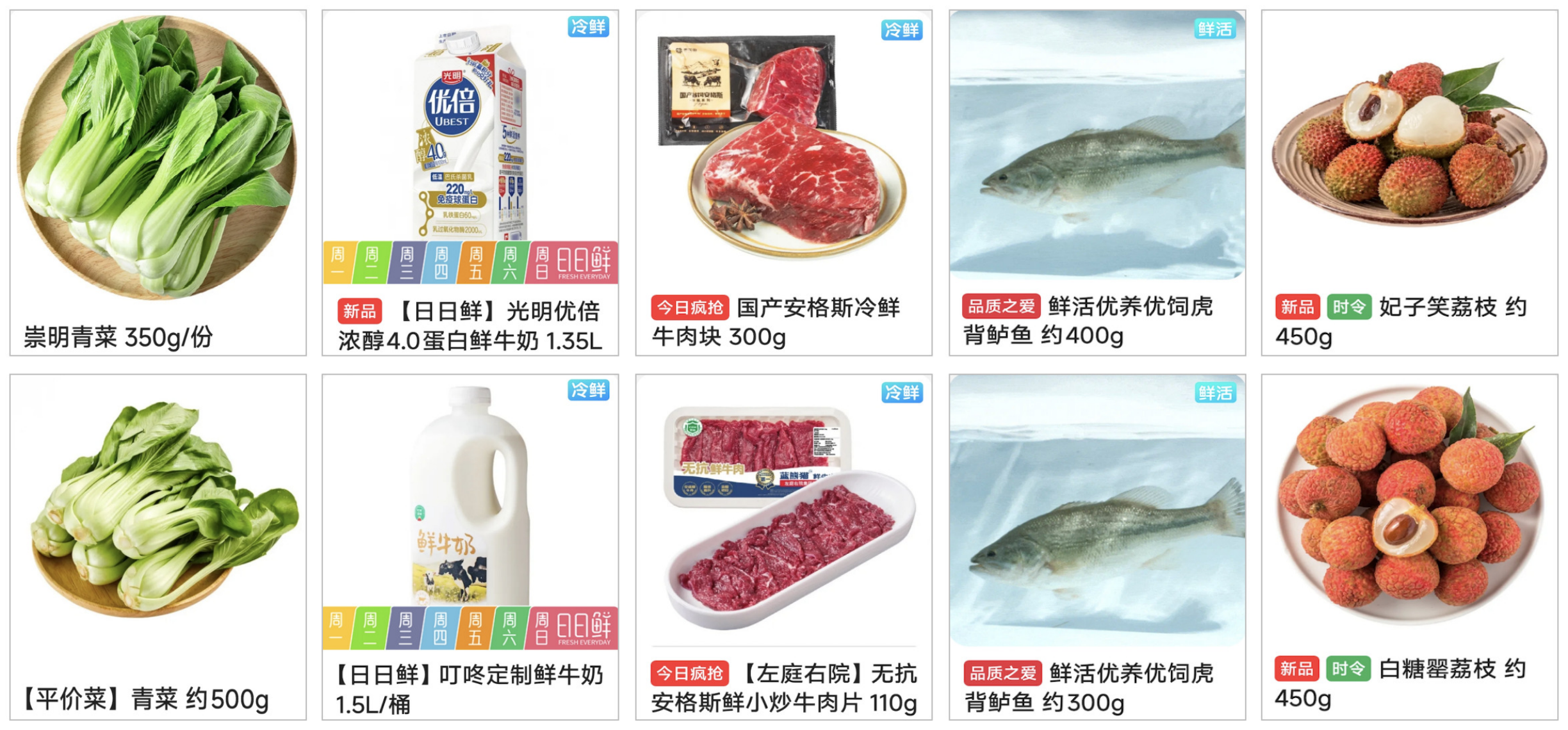} % 图片宽度适配minipage
        \caption{Various assortments of highly substitutable perishable goods. Each row presents multiple categories including vegetables, dairy (e.g., milk), meat (e.g., beef), fish, and fruits (e.g., lychees). Meanwhile, each column features similar items, such as different portions of green vegetables, milk products, beef cuts, fish, and lychee varieties, illustrating substitutable relationships within each column.
}
        \label{fig:products} % 唯一label
\end{figure}

Inventory management for perishable goods represents a critical challenge in modern retail. Suboptimal stocking decisions, compounded by product perishability and volatile demand, incur amplified financial losses from spoilage or lost sales. A persistent yet under-addressed issue stems from censored demand: the systematic bias introduced when stockout events artificially truncate observable sales data. For example, a store experiencing an 11 AM stockout of fresh strawberries may record only 100 units sold despite genuine demand for 200 units. Such discrepancies create a self-reinforcing cycle: forecasting models trained on censored sales underestimate true demand, perpetuating stockouts and waste. The problem escalates in complexity within our frontier warehouse model for fresh e-commerce. Characterized by a hyperlocal 3-kilometer service radius that enables 30-minute delivery, this model simultaneously manages assortments of highly substitutable perishables as illustrated in Figure \ref{fig:products}. These operational advantages, however, come with inherent trade-offs: the restricted coverage area exacerbates demand volatility, while the short shelf lives of products, such as seafood and fresh vegetables, create acute inventory pressure. Consequently, the combined effect of three key factors, including a limited customer base, rapid perishability rates, and high product substitutability, frequently triggers stockout events that systematically distort demand visibility.

Traditional demand forecasting methods, from ARIMA\cite{arima} to deep learning\cite{informer}\cite{itransformer}\cite{logtrans}\cite{autoformer}\cite{fedformer}\cite{pyraformer} architectures, assume that sales data accurately reflect true demand. However, this assumption collapses in the event of product stockouts, as sales volumes under-represent the latent demand. Although recent advances in temporal hierarchical forecasting\cite{athanasopoulos2017forecasting}\cite{punia2020cross}\cite{rangapuram2023coherent} and covariate integration have improved accuracy under normal conditions, and there have also been some studies on censored demand forecasting methods\cite{amjad2017censored}\cite{pedregal2024censoreddataforecastingapplying}\cite{mersereau2015demand}\cite{migueis2022reducing}\cite{sousa2025predicting}\cite{tong2018behavioral}, however, two systemic limitations in existing datasets persist:(1) the absence of explicit stockout annotations prevents distinguishing true zero demand from zero demand caused by supply shortage, and (2) daily or weekly data aggregation obscures intraday demand patterns critical for perishable replenishment, such as surges in purchases during morning or event peaks. These gaps hinder both algorithmic innovation and practical deployment.

To address these challenges, we introduce FreshRetailNet-50K, the first open-source benchmark in fresh retail through three key contributions:
\begin{itemize}
\item \textbf{Verified Stockout Event Annotations}
FreshRetailNet-50K provides the first comprehensive set of verified stockout labels at hourly resolution, enabling researchers to directly distinguish between true zero-demand periods and zero demand caused by inventory depletion. These annotations support novel approaches for reconstructing latent demand during stockout periods and training censorship-aware forecasting models - capabilities impossible with existing datasets that require indirect stockout inference from inventory gaps.

\item \textbf{Hourly Demand Dynamics}
Capturing sales at hourly resolution across 863 SKUs and 898 stores in 18 cities, the dataset reveals critical intraday patterns completely obscured in daily-aggregated data. Researchers can now observe and model morning/evening purchase surges for perishables which is essential for optimizing time-sensitive replenishment of perishable goods.

\item \textbf{Multi-Model Covariates}
The benchmark uniquely combines hyperlocal weather streams, holiday effects, and promotional discounts. This rich contextual layer supports research beyond inventory management, including climate-resilient supply chain planning and dynamic pricing\cite{censor_pricing} under perishability constraints. The dataset’s multi-city coverage further enables transfer learning for demand shocks caused by extreme weather events.
\end{itemize}

FreshRetailNet-50K pioneers the first perishable retail benchmark that incorporates stockout-induced censoring, thereby standardizing demand modeling for short-shelf-life goods. Its open-source hourly resolution data facilitates groundbreaking research in censorship-aware demand modeling, offering new opportunities for both academic studies and practical applications in the industry.

\section{Related Work}
\subsection{Retail Forecasting Datasets}
Existing public datasets for retail demand forecasting exhibit critical gaps when applied to demand forecasting of perishable products, particularly in two dimensions:

\textbf{Censoring Indicators}: 
Existing retail benchmarks (e.g., M5\cite{m5-forecasting-accuracy}\cite{makridakis2022m5}, Walmart\cite{walmart-recruiting-store-sales-forecasting}, Favorita\cite{favorita-grocery-sales-forecasting}) lack mechanisms to distinguish true zero demand from unobserved demand during stockouts, as they universally treat zero sales records as demand absence. This limitation is especially severe for perishable goods, where short shelf lives and frequent restocking requirements heighten stockout risks, leading to unrecorded genuine demand. The absence of stockout annotations forces models to process censored demand signals as accurate observations, creating systemic underestimation biases. These inaccuracies compound in hierarchical forecasting structures, where store-level stockouts distort aggregated demand patterns across temporal and spatial dimensions, ultimately compromising forecast accuracy at higher organizational levels.
This limitation triggers a vicious cycle: demand underestimation leads to poor inventory decisions, increasing stockouts and further corrupting demand measurements—a self-reinforcing deterioration known in supply chain theory as a non-stationary Markov process\cite{nasr2018markov}\cite{bayraktar2010markov}. Despite its operational impact, to the best of our knowledge, no existing public dataset provides verifiable stockout annotations, leaving a critical gap in perishable retail forecasting research. 

\textbf{Temporal Granularity}: 
Recent advances in supply chain management have increasingly emphasized the critical role of temporal granularity in demand forecasting and inventory optimization. Existing public datasets, such as M5\cite{m5-forecasting-accuracy}\cite{makridakis2022m5} and Corporación Favorita\cite{favorita-grocery-sales-forecasting}, predominantly aggregate sales data at daily or weekly intervals. While these datasets have facilitated foundational research in retail forecasting, their coarse temporal resolution limits their applicability to perishable goods management, where intra-day demand volatility and stockout dynamics significantly impact operational decisions. Coarse-grained data such as weekly aggregates obscure stockouts due to temporal averaging effects. A single day's stockout typically reduces weekly sales by just 10\%, making both detection and annotation unreliable. In contrast, hourly resolution data captures clearer zero-demand signals during stockout periods, as it reveals far more pronounced demand suppression while preserving normal demand patterns. This enables accurate stockout identification and subsequent recovery of censored demand.
Although transaction-level datasets like Olist\cite{olist_andr__sionek_2018} offer finer granularity, their design prioritizes applications in customer behavior analysis rather than supply chain optimization. This misalignment underscores a critical gap: the absence of high-resolution, public datasets tailored to perishable goods. Addressing this gap could enable novel methodologies for real-time inventory replenishment and censored demand recovery, advancing both algorithmic research and practical implementations in modern retail ecosystems.

\subsection{The Synthetic Gap in Perishable Retail}
Recent time series imputation methods like GRUI\cite{grui}, GRU-D\cite{GRU-D}, SAITS\cite{saits}, ImputeFormer\cite{imputeformer}, and diffusion-based\cite{csdi}\cite{pristi}\cite{score-cdm}\cite{sasdim} models, along with demand forecasting approaches such as TFT(Temporal Fusion Transformer)\cite{tft}, statistical models (e.g., ARIMA), and rule-based methods, are typically evaluated on standard benchmarks such as ETT\cite{informer} or Electricity\cite{electricity_dataset} datasets. While these methods are valuable for algorithmic development, they often operate in silos, imputation methods focus on filling missing data, whereas demand forecasting methods aim to predict future values. However, in the context of perishable retail, both face challenges from real-world retail censorship. Stockouts create fundamentally different missing mechanisms where demand surges during promotions generate non-random censorship patterns. Meanwhile, perishability leads to permanent demand loss when products expire, creating an irreversible data generation process that violates the missing-at-random\cite{rubin1976missing}\cite{little2019missing}\cite{little2021missing} assumption underlying most imputation and many forecasting methods. FreshRetailNet-50K addresses this limitation by providing large-scale naturally censored retail data with annotated stockout events, enabling rigorous evaluation of both imputation and demand forecasting methods against realistic supply-demand interactions. This facilitates the development of specialized approaches that can account for retail specific censorship mechanisms, fostering a more integrated approach where imputation can provide cleaner data for forecasting.

\section{Dataset Collection and Characteristics}
\subsection{Data Provenance and Integration}
FreshRetailNet-50K‌ integrates multi-source operational data from ‌898 stores‌ across ‌18 major Chinese cities‌ (e.g., Beijing, Shanghai, Guangzhou) during a ‌three-month observation period‌ (March to June 2024). Inventory dynamics are captured via warehouse management systems, tracking hourly stock levels for ‌863 perishable product categories‌, including vegetables, meat \& poultry \& eggs, fruits, seafood, frozen goods and etc. Synchronized with inventory updates, ‌hourly sales data‌ are derived from ‌online order transaction systems‌, aligning stock depletion with latent demand through granular consumer purchase records. Marketing campaigns are annotated with explicit discount rate ranges (e.g., 15\%–30\% price reductions) to establish causal anchors for demand fluctuation analysis.
To quantify exogenous influences, the dataset incorporates temporal and environmental covariates: ‌Daily meteorological data‌ (temperature, precipitation) are collected via API for each store location, while ‌systematically labeled Chinese statutory holidays‌ (e.g., Qingming Festival, Labour Day, Dragon Boat Festival) are integrated to disentangle climate sensitivity and festival-driven impacts on perishable product demand and spoilage dynamics.

To ensure compliance with data privacy regulations, all personally identifiable information was pseudonymized prior to release, with store-level identifiers encrypted to prevent reidentification. 

\subsection{Temporal Patterns and Statistical Insights}
\begin{figure}[htbp]
    \centering
    \begin{minipage}[t]{0.48\textwidth} % 第一个图片块，占约48%宽度
        \centering
        \includegraphics[width=0.95\textwidth, height=4cm]{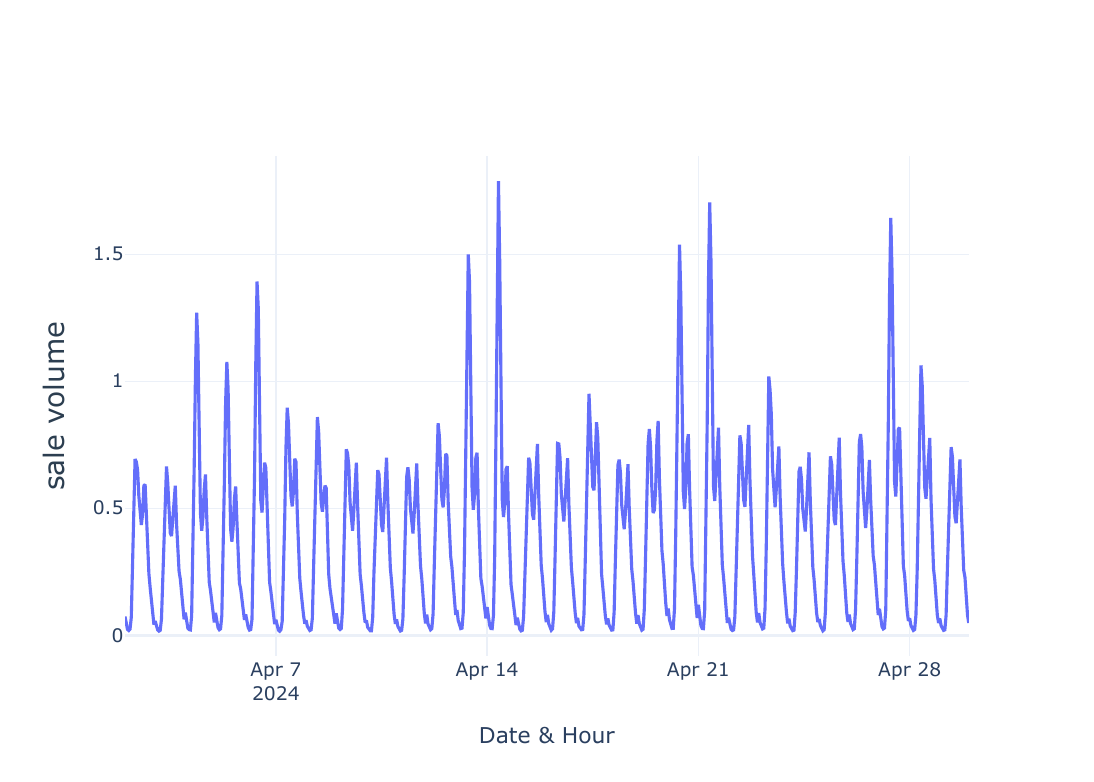} % 图片宽度适配minipage
        \caption{Temporal Demand Patterns in FreshRetailNet-50K}
        \label{fig:temporal} % 唯一label
    \end{minipage}
    \hfill % 水平间隔
    \begin{minipage}[t]{0.48\textwidth} % 第二个图片块
        \centering
        \includegraphics[width=0.95\textwidth, height=4cm]{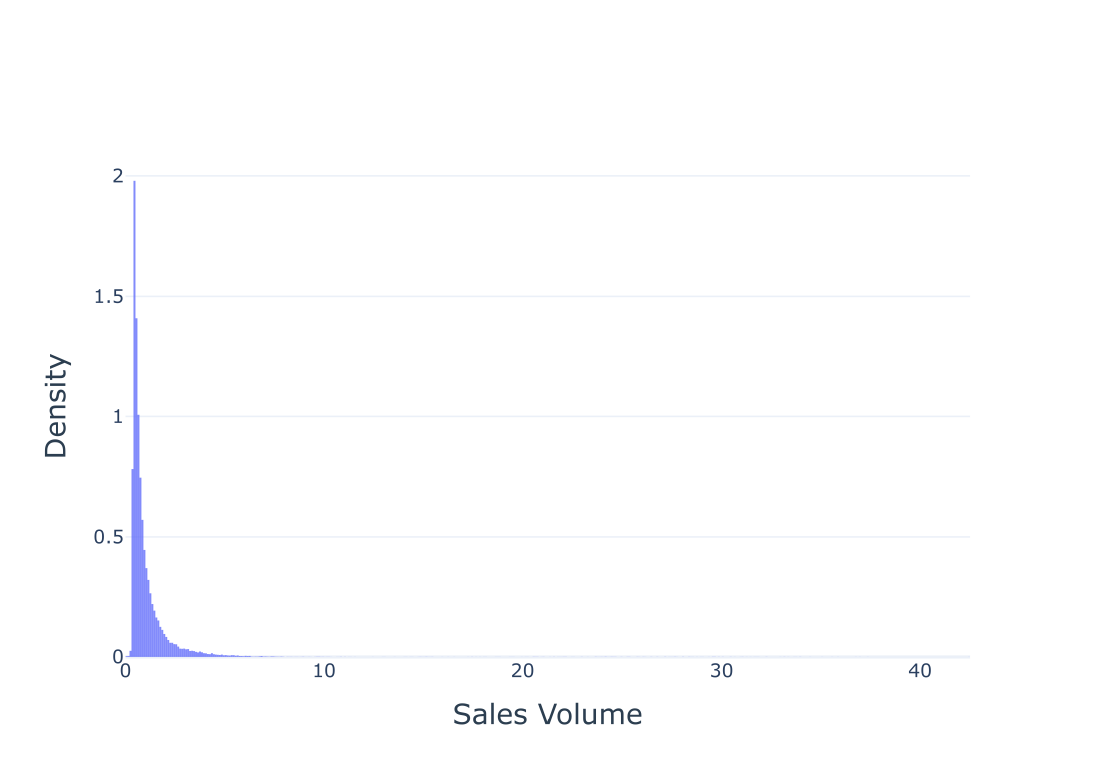}
        \caption{Demand Concentration and Long-Tail Distribution}
        \label{fig:longtail} % 唯一label
    \end{minipage}
\end{figure}

‌The FreshRetailNet-50K dataset systematically characterizes perishable retail demand through three interconnected complexity dimensions. First, temporal cyclicity exhibits distinct intraday regimes: weekday demand shows bimodal peaks at 9:00 AM for pre-lunch preparation and 4:00 PM for pre-dinner shopping, while weekends feature a single amplified morning surge at 9:00 AM with doubled peak intensity, reflecting household bulk purchasing behaviors as shown in Figure \ref{fig:temporal}.
Second, demand heterogeneity follows a power-law distribution\cite{clauset2009power} ($\alpha=2.83$, Kolmogorov-Smirnov\cite{berger2014kolmogorov} $p<0.03$), where 20\% of SKUs account for 51.8\% of transactions (Figure \ref{fig:longtail}). This creates operational duality: high-velocity staples dominate revenue streams while long-tail perishables exhibit sparse yet critical demand patterns that complicate low-frequency item forecasting.
Third, exogenous factors demonstrate measurable impacts: promotional campaigns yield median sales uplifts of $1.43\times$ with interquartile range $1.05\times$ -- $2.06\times$ after controlling for baseline 27\% holiday surges. Environmental analysis reveals an 11\% rainfall-correlated increase in vegetable demand, likely through outdoor market substitution. 
By integrating temporal, product-hierarchy, and environmental dimensions, FreshRetailNet-50K advances perishable retail benchmarking beyond isolated demand prediction to holistic inventory optimization under promotional, climatic, and operational perturbations. This integration supports causal analysis of perishability risks while preserving real-world supply chain dynamics through rigorously aligned observational data.

\subsection{Covariate-Dependent Censoring Mechanisms}
Stockout events in FreshRetailNet-50K exhibit structured missingness patterns that fundamentally challenge conventional missing-at-random\cite{rubin1976missing}\cite{little2019missing}\cite{little2021missing} assumptions. A critical observation is the inherent temporal dependency of stockout rates, which follow a pronounced U-shaped diurnal cycle. Specifically, restocking operations at 6 AM usually reduce stockout rates to below 2\%, after which inventory attrition progresses monotonically, culminating in peak stockout probabilities of 26\% by 8 PM. This cyclical pattern persists until the next replenishment cycle, creating time-dependent censoring where late-day transactions systematically underrepresent true demand due to depleted inventories.
Promotional depth and environmental factors further drive non-random censoring. Quantile-binned (model specification: $Stockout=\beta\cdot Rank(X)+\epsilon$, where $Rank$ maps covariates to bins with ascending ranks) analysis demonstrates that deeper price reductions paradoxically increases stockouts($\beta = -0.046$, $p = 0.018$) by amplifying demand beyond inventory capacity, creating a self-reinforcing cycle where temporary demand surges exacerbate inventory depletion, a phenomenon that synthetic datasets fail to capture. Rainfall elevates stockouts through offline-to-online demand migration ($\beta=0.108$, $p=0.05$), while temperature exhibits category-divergent effects: rising temperatures increase frozen goods stockouts ($\beta=0.065$, $p=0.024$) but decrease meat \& poultry \& eggs shortages ($\beta=-0.067$, $p=0.033$). These dynamics necessitate modeling stockouts as multivariate interactions of demand, lagged supply, and environmental contexts, where censoring arises from structured covariate dependencies rather than random noise. FreshRetailNet-50K uniquely encodes these retail-specific mechanisms, enabling robust demand estimation under real-world inventory constraints.

\subsection{Benchmark Differentiation}
The FreshRetailNet-50K dataset fundamentally advances perishable retail forecasting through three core differentiators from existing benchmarks (Table \ref{tab:dataset_vs}). First, its hourly temporal resolution uniquely captures stockout-induced demand censoring patterns, where inventory depletion directly truncates observable sales, a critical capability absent in daily (M5\cite{m5-forecasting-accuracy}\cite{makridakis2022m5}, Favorita\cite{favorita-grocery-sales-forecasting}) or weekly (Walmart\cite{walmart-recruiting-store-sales-forecasting}) granularity. Second, the explicit stockout annotations provide natural MNAR (missing-not-at-random)\cite{rubin1976missing}\cite{little2019missing}\cite{little2021missing} censoring mechanisms, enabling rigorous evaluation of demand recovery models under real-world supply constraints. Third, the multi-level hierarchy spans both geographic (city-store) and product (category-SKU) dimensions, allowing joint analysis of regional perishability risks and item-level replenishment dynamics.

While the series count (50K) is smaller than Favorita’s 174K, the dataset matches its 0.1B sample scale through fine-grained hourly observations, each integrating censored demand signals with perishability-specific covariates. By unifying high-frequency censoring, hierarchical contexts, and demand-inventory causality, FreshRetailNet-50K establishes a targeted benchmark for inventory-aware forecasting in perishable retail, addressing gaps in general-purpose retail datasets.

\begin{table}[htbp]
\centering
\caption{Comparative analysis against retail demand forecasting benchmarks}
\label{tab:dataset_vs}
\begin{tabular}{c|cccc}
    \hline 
    \textbf{Dimension} & \textbf{M5} & \textbf{Favorita} & \textbf{Walmart} & \textbf{FreshRetailNet-50K} \\
    \hline 
    Stockout Annotation & No & No & No & \textbf{Yes} \\
    Censoring Mechanism & No & No & No & \textbf{MNAR} \\
    Temporal Granularity & Daily & Daily & Weekly & \textbf{Hourly} \\
    Hierarchy Structure & \textbf{Yes} & No & No & \textbf{Yes} \\
    Series \& Samples & 30K\&59M & \textbf{174K\&0.1B} & 3K\&0.4M & \textbf{50K\&0.1B}\\
    \hline
\end{tabular}
\end{table}

\section{‌Censored Demand Modeling: Recovery and Forecasting}
The structured missingness in FreshRetailNet-50K requires rethinking conventional approaches to demand modeling. We formalize two interconnected tasks that address censoring-induced biases in both retrospective demand recovery and prospective demand forecasting.
\subsection{Task 1: Latent Demand Recovery}
\label{sec:recovery}
\textbf{Problem Formulation}

The stockout-induced censoring in FreshRetailNet-50K requires reconstructing latent demand $\mathbf{d}$ from partially observed sales $\mathbf{y}$.
Let $X\in R^{T\times F}$ denote the multivariate time series, where $T$ is the number of hourly timestamps and $F$ includes censored sales $\mathbf{y}$, promotional discounts $\mathbf{p}$, weather features $\mathbf{w}$, calendar indicators $\mathbf{c}$ and censoring indicator $\mathbf{s}$. 
Under the MNAR mechanism, the censoring indicator $\mathbf{s}$ depends on inventory levels $\mathbf{i}_t$:  
\begin{align}  
\mathbf{s}_t = \begin{cases}  
1 & \text{if } \mathbf{i}_t > 0 \text{ (stock available)} \\  
0 & \text{if } \mathbf{i}_t = 0 \text{ (stockout)}  
\end{cases}  
\end{align}  

The goal here is to design a model for the estimation of missing demand from input time series and mask data. 
Let $\hat{\mathbf{d}}$ denote the estimated missing demand, the final reconstructed latent demand $\mathbf{d}$ is given by 
\begin{align}
\label{eq:reconstruct}
\mathbf{d} &= \mathbf{y} \odot \mathbf{s} + \hat{\mathbf{d}} \odot (1-\mathbf{s})\\
\mathbf{\hat{d}}&=f_\theta(\underbrace{\mathbf{y}\odot\mathbf{s}}_{Observed\, sales}, \underbrace{\mathbf{p}, \mathbf{w}, \mathbf{c}}_{Covariate}, \underbrace{\mathbf{s}}_{Stockout})
\end{align}
where $\odot$ denotes the element-wise product. Essentially, we take estimated demand if inventory runs out and observed sales otherwise for demand recovery.

\textbf{Baseline Methods}

To assess FreshRetailNet-50K’s latent demand recovery challenges, we benchmark methods spanning four paradigms:  

\begin{itemize}

\item  Multi-Periodicity Models: TimesNet\cite{timesnet}, capturing hierarchical temporal patterns (intraday and daily cycles) critical for perishable sales temporal dynamics.  
\item  Attention-Based Imputation: SAITS\cite{saits} and ImputeFormer\cite{imputeformer} model cross-temporal dependencies in censored regions. iTransformer\cite{itransformer} enables multidimensional covariate fusion (promotions, weather) through channel-first attention\cite{attention}.
    
\item  Uncertainty-Aware Generators: GPVAE\cite{gpvae} (temporal smoothness priors\cite{gp}) and CSDI\cite{csdi} (diffusion-based denoising), addressing MNAR-induced uncertainty in stockout events.  
    
\item  Interpretable Baselines: DLinear\cite{dlinear}, decomposing demand into trend and seasonality for robustness validation.  
\end{itemize}
%This spectrum covers multi-periodicity modeling (TimesNet), covariate-aware attention (SAITS, ImputeFormer), probabilistic modeling (CSDI), and interpretable linear methods, evaluating how different inductive biases address perishable retail’s censoring dynamics.  

\subsection{Task 2: Censoring-Robust Demand Forecasting}
\label{sec:forecasting}
\textbf{Problem Formulation}

The recovered demand series $d$ from Section~\ref{sec:recovery} enables the construction of de-biased daily demand $D_t=\sum_{\tau=24k}^{24(k+1)}d_\tau$. We formulate the 7-day ahead prediction as a covariate-enhanced regression problem:
$$
\hat{D}_{T+1:T+7}=f(\underbrace{D_{T-H, T}}_{Historical\,Demand}, \underbrace{P_{T-H:T+7}, W_{T-H:T+7}, C_{T-H:T+7}}_{Coviriate})
$$
The covariates $P$, $W$, and $C$, representing promotional activities, weather conditions, and calendar events respectively, are assumed to be known for the forecast horizon $T+1$ to $T+7$. This premise holds as $P$ can be anticipated from marketing plans, $W$ can be derived from weather forecasting, while $C$ is determined by holiday schedules and weekend arrangements.

\textbf{Baseline Methods}

We evaluate three distinct forecasting paradigms to assess methodological generality:
\begin{itemize}
\item \textbf{Similar Scenario Average (SSA)} - a statistical baseline that uses weighted historical demands from similar scenarios for forecasting, where similarity is determined by the alignment of promotional and weather patterns.
\item \textbf{Temporal Fusion Transformer (TFT)}\cite{tft} - a deep learning model that explicitly models promotional impacts through covariate-gated multi-head attention for time series regression. 
\item \textbf{DLinear}\cite{dlinear} - a machine learning model that decomposes series into trend-cyclical and seasonal components via learnable moving average filters for time series regression.
\end{itemize}
To evaluate the de-censoring impact on forecasting, we contrast two paradigm classes across data pipelines:
\begin{itemize}
\item \textbf{Raw Pipeline}: Directly aggregate observed sales without Latet Demand Recovery.
\item \textbf{Imputation-Augmented}:  Utilizes daily aggregates from top-performing Censored Demand Recovery models, such as TimesNet\cite{timesnet} and ImputeFormer\cite{imputeformer}.
\end{itemize}

\section{Experimental Results and Analysis}
\subsection{Evaluation Framework}
The latent demand recovery problem is fundamentally constrained by the unobservability of true demand during stockouts. Our evaluation framework combines:

\begin{itemize}
\item \textbf{Controlled MNAR Simulation}: Using non-stockout periods as ground truth, we generate synthetic censored regions matching empirical patterns. Model performance is quantified through Weighted Absolute Percentage Error(WAPE) and Weighted Percentage Error(WPE):
\begin{align}
\text{WAPE} &= \frac{\sum_t|d_t - y_t|}{\sum_t y_t} \quad \text{(magnitude accuracy)} \\
\text{WPE} &= \frac{\sum_t(d_t - y_t)}{\sum_t y_t} \quad \text{(bias direction)}
\end{align}

\item \textbf{Demand-Supply Decoupling}: A core objective of censored demand recovery is to disentangle stockout-induced demand suppression from true consumption patterns. We quantify this decoupling through:
\begin{equation}
\rho_{DS} = \sum_{i\in \mathcal{P}} w_i \cdot Pearson(\mathbf{SR}_i, \mathbf{d_i)}, \quad w_i = \frac{\mu_i}{\sum \mu_i}
\end{equation}
where $\rho_{DS}$ denotes the Decoupling Score measuring the weighted correlation between stockout ratios and recovered demand. The set $\mathcal{P}$ contains all store-product pairs, with $\mathbf{SR}_i$ representing the stockout ratio, $\mathbf{d}_i$ the recovered demand, and $\mu_i$ the mean sales for each store--product pair $i \in \mathcal{P}$.
\end{itemize}

As demand forecasting employs time series supervised learning models, evaluation primarily relies on supervised learning metrics. To assess predictive capability for true demand, evaluations are conducted exclusively during operational periods without stockouts. Metrics focus not only on absolute error (WAPE) for accuracy but also on bias (WPE) to quantify over/under-prediction impacts on stockouts and losses.

\subsection{Latent Demand Recovery Performance}
\label{exp:recovery}
The evaluation reveals critical insights into latent demand recovery under stockout-induced MNAR mechanisms. TimesNet\cite{timesnet} achieves state-of-the-art performance with 27.62\% WAPE and 1.43\% WPE, demonstrating both high magnitude accuracy and minimal systematic bias. This aligns with its multi-periodicity modeling capability, effectively capturing perishable sales’ intraday cycles (9AM/4PM peaks) and interday cycles(weekend surges) while mitigating censoring distortions. DLinear\cite{dlinear} demonstrates competitive WAPE (29.99\%) and bias (-12.85\%), validating linear models’ utility as robust baselines for perishable demand recovery. Notably, transformer-based methods show divergent behaviors: ImputeFormer\cite{imputeformer} underperforms iTransformer\cite{itransformer}, likely due to the latter’s channel-first attention\cite{attention} better fusing promotions and weather covariates. SAITS’\cite{saits} higher error likely aligns with two dataset-specific challenges: long-range and multi-periodicity demand dependenct and hourly zero-inflation distribution, which may destabilize standard self-attention.

\begin{table*}[htbp]
  \caption{Performance comparison of models on MNAR}
  \label{tbl:overall_res_mnar}
  \resizebox{\linewidth}{!}{ % 保持宽度适应文本宽度
  \begin{tabular*}{1.15\linewidth}{@{\extracolsep{\fill}} c|cccccccc}
  \hline
  Metric & Raw\, Sale & TimesNet & ImputeFormer & SAITS & iTransformer & GPVAE & CSDI &DLinear\\\hline
  WAPE & - & \textbf{27.62\%} & 35.65\% & 61.54\% & 33.30\% & 56.08\% & 39.54\% & 29.99\% \\
  WPE & - & \textbf{1.43\%} & -26.16\% & -59.74\% & -20.95\%  & 12.02\% & -23.48\% & -12.85\% \\
  $\mathbf{\rho_{DS}}$ & -0.57 & \textbf{0.07} & -0.26 & -0.50 & --0.22  & -0.15 & -0.27 & -0.16 \\
  \hline
  \end{tabular*}
  }
  \end{table*}

The Decoupling Score ($\rho_{DS}$) further validates TimesNet’s superiority: its near-zero correlation (0.07 vs. raw sales' -0.57) indicates successful elimination of spurious demand and stockout linkages. Comparatively, models such as SAITS and CSDI still maintain significant negative correlations, reflecting that their recovered demand is still under the implicit influence of supply constraints. Notably, although generative model GPVAE performs moderately on the decoupling metric ($-0.15$), its 55.36\% WAPE reveals that the probabilistic generation method may over-smooth fluctuations in actual demand in fresh produce scenarios. 

These results underscore FreshRetailNet-50K's value in benchmarking both accuracy (WAPE/WPE) and causal validity ($\rho_{DS}$)—a dual requirement for perishable retail where biased demand estimates directly propagate to stockout cascades. The dataset’s hourly censoring annotations enable fine-grained analysis of temporal recovery patterns, a capability absent in coarse-grained benchmarks.

\subsection{Forecasting Performance with Recovered Demand}
To validate the impact of demand recovery on forecasting reliability, we compared model performances across latent demand recovery methods (Table \ref{tab:forecasting_comparison}). For the TFT on the full dataset, demand recovery via both TimesNet and iTransformer reduced WAPE from 31.75\% (raw sales) to 29.02\% and 29.26\%, respectively. This improvement was accompanied by a mitigation of systematic underestimation: WPE improved from -7.37\% to 2.58\% with TimesNet and to near-neutral 0.57\% with iTransformer, balancing accuracy and bias correction.

\begin{table*}[htbp]
  \centering
  \caption{Performance comparison of models on different latent demand recovery methods}
  \label{tab:forecasting_comparison}
  \resizebox{\linewidth}{!}{ 
    \begin{tabular}{c c c c c c c c} 
      \toprule
      \multirow{2}{*}{\textbf{Group}} & \multirow{2}{*}{\textbf{\makecell{Recovery \\ Method}}} 
      & \multicolumn{2}{c}{\textbf{SSA}} & \multicolumn{2}{c}{\textbf{TFT}} & \multicolumn{2}{c}{\textbf{DLinear}} \\
      \cmidrule(lr){3-4} \cmidrule(lr){5-6} \cmidrule(lr){7-8} 
      &  & \multicolumn{1}{c}{WAPE} & \multicolumn{1}{c}{WPE} & \multicolumn{1}{c}{WAPE} & \multicolumn{1}{c}{WPE} & \multicolumn{1}{c}{WAPE} & \multicolumn{1}{c}{WPE} \\
      \midrule
      % Overall 分组：WAPE取最小，WPE取绝对值最小
      \multirow{5}{*}{Overall} 
        & TimesNet       & 29.79\%  & \textbf{1.39\%}   & \textbf{29.02\%}  & 2.58\%   & 30.73\%  & 6.88\%  \\  % SSA WPE绝对值最小（1.39%）；TFT WAPE最小（29.02%）
        & ImputeFormer   & 29.80\%  & -4.19\%  & 29.54\%  & 2.77\%   & \textbf{30.27\%}  & \textbf{1.91\%}   \\  % DLinear WAPE最小（30.27%）；DLinear WPE绝对值最小（1.91%）
        & iTransformer   & \textbf{29.76\%}  & -2.12\%  & 29.26\%  & \textbf{0.57\%}   & 30.69\%  & 4.35\%   \\  % SSA WAPE最小（29.76%）；TFT WPE绝对值最小（0.57%）
        & DLinear        & 30.00\%  & -1.49\%  & 29.71\%  & 2.63\%   & 30.60\%  & 4.02\%   \\
        & Raw Sales      & 31.97\%  & -10.50\% & 31.75\%  & -7.37\%   & 31.56\%  & -4.89\% \\
      \midrule
      % Low Sale 分组：WAPE取最小，WPE取绝对值最小
      \multirow{5}{*}{Low Sale} 
        & TimesNet       & 37.87\%  & 4.89\%   & 37.33\%  & 7.78\%   & 40.73\%  & 12.37\%   \\
        & ImputeFormer   & 37.66\%  & -4.77\%   & 37.51\%  & 2.55\%   & \textbf{39.75\%}  & 3.05\%   \\  % DLinear WAPE最小（39.75%）
        & iTransformer   & \textbf{37.63\%}  & \textbf{-1.75\%}   & \textbf{37.04\%}  & \textbf{0.17\%}   & 40.33\%  & 6.73\%   \\  % SSA WAPE最小（37.63%）；SSA WPE绝对值最小（1.75%）；TFT WAPE最小（37.04%）；TFT WPE绝对值最小（0.17%）
        & DLinear        & 37.74\%  & 1.89\%   & 38.09\%  & 9.16\%   & 40.35\%  & 9.16\%   \\
        & Raw Sales      & 38.33\%  & -8.03\%   & \textbf{37.04\%}  & -1.72\%   & 39.78\%  & \textbf{-1.03\%}   \\  % TFT WAPE并列最小（37.04%）；DLinear WPE绝对值最小（1.03%）
      \midrule
      % High Sale 分组：WAPE取最小，WPE取绝对值最小
      \multirow{5}{*}{High Sale} 
        & TimesNet       & \textbf{23.98\%}  & \textbf{-1.12\%}   & \textbf{23.03\%}  & \textbf{0.87\%}   & 23.58\%  & 2.98\%   \\  % SSA WAPE最小（23.98%）；SSA WPE绝对值最小（1.12%）；TFT WAPE最小（23.03%）；TFT WPE绝对值最小（0.87%）
        & ImputeFormer   & 24.15\%  & -3.76\%   & 23.82\%  & 1.21\%   & \textbf{23.47\%}  & 1.13\%   \\  % DLinear WAPE最小（23.47%）
        & iTransformer   & 24.10\%  & -2.36\%   & 23.66\%  & \textbf{0.87\%}   & 23.78\%  & 2.69\%   \\  % TFT WPE并列最小（0.87%）
        & DLinear        & 24.43\%  & -2.36\%   & 23.68\%  & -2.05\%   & 23.61\%  & \textbf{0.36\%}   \\  % DLinear WPE绝对值最小（0.36%）
        & Raw Sales      & 27.40\%  & -12.26\%   & 27.95\%  & -11.46\%   & 25.68\%  & -7.63\%   \\
      \bottomrule
    \end{tabular}
  }
\end{table*}

\begin{figure}[htbp]
    \centering
    \begin{minipage}[t]{0.48\textwidth} % 第一个图片块，占约48%宽度
        \centering
        \includegraphics[width=0.95\textwidth, height=0.5\textwidth]{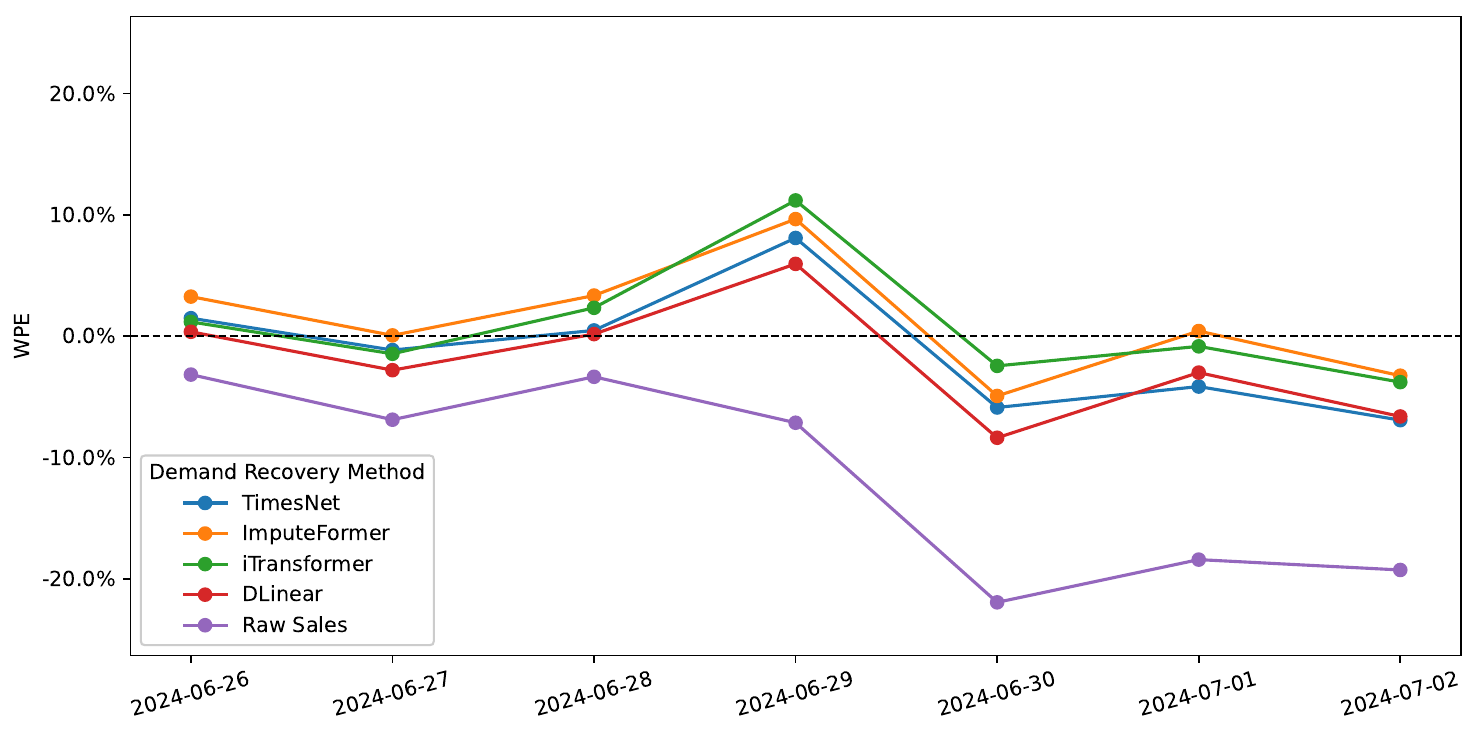} % 图片宽度适配minipage
        \caption{WPE Over Time by Different Demand Recovery Method On High Sale}
        \label{fig:wpe_hight} % 唯一label
    \end{minipage}
    \hfill % 水平间隔
    \begin{minipage}[t]{0.48\textwidth} % 第二个图片块
        \centering
        \includegraphics[width=0.95\textwidth, height=0.5\textwidth]{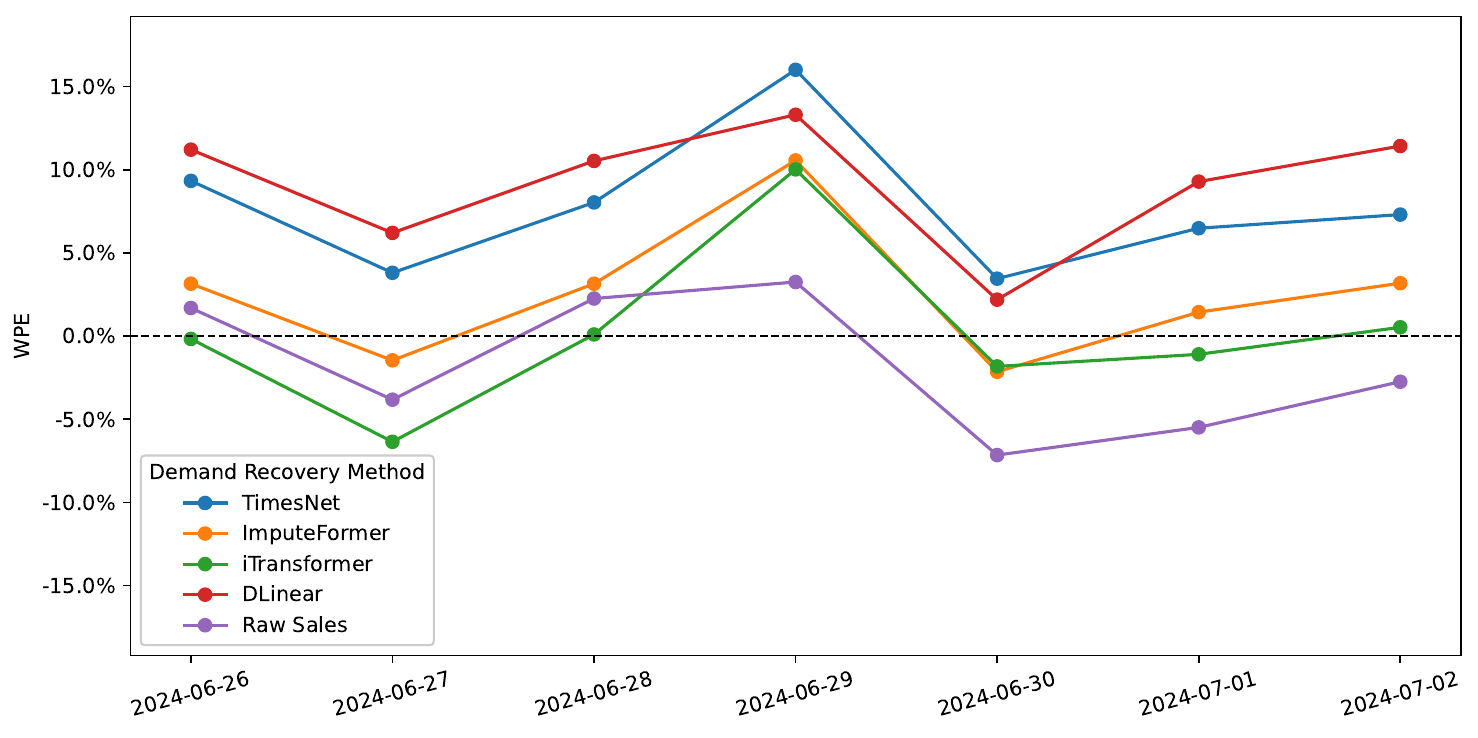}
        \caption{WPE Over Time by Different Demand Recovery Method On Low Sale} 
        \label{fig:wpe_low} % 唯一label
    \end{minipage}
\end{figure}

In high sale group, TFT demonstrated superior adaptability, achieving near-neutral bias: WPE values were 0.87\% via TimesNet and iTransformer. This stands in stark contrast to the -11.46\% underestimation bias observed in raw data, with corresponding WAPE improvements of 4.95\% and 4.29\%. As illustrated in Figure \ref{fig:wpe_hight}, which depicts WPE over time in the high sale group, various filling methods notably improved WPE, keeping it near 0. Statistical models such as SSA improved high sale WPE from -12.26\% to -1.12\% with TimesNet recovery but lagged in accuracy, highlighting their limited capacity to leverage temporal patterns. DLinear showed competitiveness in high sale scenarios, though its performance diverged significantly in low sale contexts. Low sale items’ inherent sparsity and volatility presented critical challenges. While SSA modestly reduced WPE from -8.03\% to -1.75\%, TFT exhibited minimal bias improvement, and DLinear shifted to severe overestimation, becoming the worst performer among all models. As shown in Figure \ref{fig:wpe_low}, for the low sale group, most filled results showed overestimation, aligning with the described challenges. This divergence underscores how sparse, noisy data distributions undermine model generalization and expose vulnerabilities in demand recovery and forecasting for low volatility SKUs.

Overall, demand recovery systematically alleviated underestimation through reduced negative WPE values and improved accuracy through lowered WAPE. TFT stood out for its robust adaptability across all dataset segments, consistently outperforming other models. DLinear’s mixed performance—strength in high-sale scenarios versus failure in low-sale contexts—highlighted method-specific limitations tied to data density.  

FreshRetailNet-50K establishes a new paradigm by linking demand recovery quality to forecasting reliability. Success is no longer measured solely by point prediction accuracy but by the ability to disrupt the stockout-forecast bias cycle. Grounded in real-world operational constraints, these findings position FreshRetailNet-50K as an indispensable benchmark for developing inventory solutions that reconcile machine learning innovation with the unforgiving realities of perishable retail.

\section{Limitation}
FreshRetailNet-50K provides valuable insights into perishable retail forecasting but should be contextualized with considerations. The temporal scope of the dataset may not fully capture seasonal trends or long-term shifts in consumer behavior and supply chain dynamics relevant to fresh produce retail. Furthermore, the sparse and volatile data of low-sale items pose challenges to model stability, as existing methods face difficulties in managing long-tail retail SKUs effectively. Lastly, the study focuses on a two-stage demand recovery-forecasting framework and does not explicitly evaluate end-to-end models, which could limit joint optimization of bias correction and prediction accuracy. These aspects highlight potential avenues for future research, such as dynamic data expansion, sparse data modeling, and the development of integrated forecasting frameworks.

\section{Conclusion}
FreshRetailNet-50K pioneers a critical advance in perishable supply chain research by providing the first open benchmark that captures the intricate interplay between demand censoring, temporal granularity, and real-world operational constraints. Through its unique integration of hourly sales trajectories, verified stockout labels, and multi-modal covariates, the dataset exposes fundamental limitations in existing demand recovery and forecasting methods. Our analysis reveals that even state-of-the-art models struggle to disentangle stockout-induced demand suppression from true consumption patterns, particularly for long-tail SKUs with sparse transactions. Released under CC-BY-4.0, FreshRetailNet-50K empowers both academia and industry to tackle pressing challenges in perishabble supply chains. Researchers may leverage its covariates to model promotion-weather interactions or design sparsity-aware architectures, while practitioners can stress-test inventory policies against demand shocks mirrored from real retail operations. By bridging machine learning with supply chain pragmatism, the dataset reorients the field toward a new paradigm: one where success is measured not merely by predicting demand, but by enabling censoring-robust replenishment that dynamically adapts to perishability constraints and market volatility.

\newpage
\bibliographystyle{plain} % 指定参考文献样式（如 plain, apa, ieee 等）
\bibliography{references} % 指定 .bib 文件名（此处为 references.bib）

\end{document}